\PassOptionsToPackage{numbers, sort&compress}{natbib}
\documentclass{article}

\usepackage[preprint]{neurips_2026}

\usepackage[utf8]{inputenc}
\usepackage[T1]{fontenc}
\usepackage{hyperref}
\usepackage{url}
\usepackage{booktabs}
\usepackage{amsfonts}
\usepackage{amsmath}
\usepackage{nicefrac}
\usepackage{microtype}
\usepackage{xcolor}
\usepackage{graphicx}
\usepackage{longtable}
\usepackage{placeins} 
\usepackage{float}

\title{The Riddle Riddle: Testing Flexible Reasoning in Large Language Models and Humans}

\author{
  Bella Fascendini\textsuperscript{1} \thanks{Code and data available at https://github.com/bellafascendini/riddle-riddle.}
  \quad
  Kathryn McGregor\textsuperscript{1}
  \quad
  Max D.~Gupta\textsuperscript{2}
  \quad
  Thomas L.~Griffiths\textsuperscript{1,2}
  \\[0.5em]
  \textsuperscript{1}Department of Psychology, Princeton University \quad
  \textsuperscript{2}Department of Computer Science, Princeton University
}

\begin{document}

\maketitle

\begin{abstract}
Humans flexibly adapt their reasoning strategies to the requirements of a given problem. Large language models (LLMs) have performed well on many cognitive tasks, however, it is unclear whether accuracy on these tasks is a result of pattern matching from training data or flexible reasoning. Here, we introduce a novel paradigm to test this question---the riddle riddle paradigm. Riddle riddles are word problems written to mimic popular riddles, but they have been altered so their answers only require literal interpretations of the problem. Identifying correct answers in this paradigm requires subjects to look past the structure of each question and flexibly apply different reasoning strategies based on the content. If LLMs are responding to surface features, such as form, a riddle-like structure should cause models to use an inventive reasoning strategy even when a literal interpretation would suffice. Alternatively, if LLMs are reasoning based on content, they should be able to flexibly apply different reasoning strategies when appropriate. Across two experiments with nine state-of-the-art LLMs and 100 human participants, we show humans and LLMs fail on this paradigm in opposite directions. LLMs were far more accurate on genuine riddles than on riddle riddles (84.9\% vs.\ 50.7\%); whereas humans showed the reverse effect (50.5\% vs.\ 80.5\%). Additionally, error analysis shows that 90.8\% of LLM errors on riddle riddles (the condition where they show diminished performance) were due to inappropriate use of inventive reasoning while only 57.6\% of human errors on genuine riddles were due to overextending literal reasoning. This error analysis shows that while both groups (humans and LLMs) make mistakes, reasoning mistakes are made more often by LLMs than by humans. Overall, LLMs' strong performance on genuine riddles may reflect retrieval of memory rather than flexible strategy selection, and without stimuli designed to elicit this contrast, it becomes easy to conflate LLM generated outputs that look like reasoning with genuine reasoning about question content. 

\end{abstract}

\section{Introduction}

Large language models (LLMs) have demonstrated remarkable performance across a wide range of reasoning tasks, from mathematical problem-solving to commonsense inference \cite{wei2022chain, lewkowycz2022solving, brown2020language, bubeck2023sparks, zellers2019hellaswag}. Performance on benchmark tasks has increasingly been taken as evidence that LLMs can achieve human-like reasoning ability, but performance alone is insufficient to support this claim. \citet{newell1972human} argued that demonstration of correct outputs---what they referred to as sufficiency proof---is only the first step toward understanding intelligent behavior; it shows that a system can produce the right answer, but doesn't tell us much about the process that generated it. Models may achieve high performance by matching surface features of problems to responses seen during training, rather than evaluating what each problem actually requires. Teasing apart these two accounts requires a paradigm that can separate what a problem looks like from what it requires.  

Riddles are a classic and particularly useful reasoning benchmark. Solving riddles requires paying attention to the structure carefully, reasoning about the alternative meanings of each word, and identifying the trick before finally arriving at a non-obvious or inventive interpretation. However, riddles are also widely distributed across the internet and in popular culture, and most importantly, so are the solutions and reasoning behind each riddle. As reasoning itself becomes part of the training data for LLMs, it becomes increasingly difficult to distinguish a model that reasons from one that has memorized what reasoning \textit{looks} like. Thus, accuracy on riddles cannot tell us whether a model is reasoning or retrieving answer from memory. One approach for testing this question is to remove the trick from the riddle entirely, which creates problems that look like riddles but only require literal interpretation and are not readily available on the internet. 

To illustrate this point, consider the following riddle: ``A cowboy rides into town on Friday, stays for three days, and rides out on Friday. How is that possible?'' To arrive at the correct answer to this riddle, one must reason creatively about the problem. It is intuitive to think of Friday as a day of the week, but this riddle requires one to interpret “Friday” as the horse’s name. Now consider a modified version of this riddle: ``A cowboy rides into town on Friday, stays for three days, and rides out on Monday. How is that possible?'' This new version preserves the original riddle’s structure but removes the requirement for inventive reasoning. Instead of having to realize the trick of considering Friday as the name of the horse, one can just take Friday to mean the day of the week. Simply put, the answer to the variant, which we call a riddle riddle, is that Friday plus three days is Monday. Riddle riddles can be defined as questions that look like popular riddles but only require straightforward interpretation and literal reasoning to reach a correct answer. 

These two variants are structurally identical, but they differ in whether the problem contains a trick. A system that selects strategies based on surface features would apply inventive reasoning to both problem types equally; however, a solver that reasons about the problem would recognize that the riddle riddle only requires literal interpretation and switch strategies accordingly. The capacity to select strategies based on what a problem requires, rather than how it appears, is a core aspect of flexible reasoning \citep{flavell1979metacognition}---inventive reasoning is very useful for solving genuine puzzles but wastes both cognitive and computational resources and could introduce potential errors on simpler problems that only require literal interpretations. Critically, this flexibility in reasoning is not trivial---it represents a genuine cognitive challenge. For example, in human development, children often over-apply newly learned strategies before learning when each is appropriate \cite{siegler1996emerging}. 

The riddle riddle paradigm thus allows us to directly test whether models evaluate what a problem \textit{requires} or whether they simply respond to how it \textit{looks} (e.g., overgeneralize answers that require inventive reasoning to any word problem that is similar to a riddle). If models rely on such heuristics, the riddle formatting alone should trigger inventive reasoning even when a literal interpretation is simpler and sufficient. Recent work has documented a similar phenomenon (``the illusion illusion'') in vision language models, which respond to illusion-like visual stimuli regardless of whether any illusion is actually present \cite{ullman2024illusion, sclar2024quantifying, goodfellow2015explaining}.

In this paper, we use the riddle riddle paradigm to test flexible reasoning across two experiments. Experiment 1 evaluates nine state-of-the-art LLMs (including GPT-5.4, Claude-opus-4-6 \& Gemini-3.1-Pro-preview) on 30 riddle sets. Experiment 2 tests human adults on the same riddles. As expected, LLMs perform significantly better on genuine riddles, compared to humans. On riddle riddles, however, humans outperform LLMs. This reversed pattern suggests that the high model performance on genuine riddles may reflect knowledge retrieval rather than flexible reasoning: when the problem is truly out of distribution and could not have been encountered during training, model performance drops. Together, these results shed light on how humans and LLMs select reasoning strategies, and where they differ.

\section{Background}
To understand whether state-of-the-art LLMs have achieved flexible reasoning, we first examine what this capacity entails in humans and where previous research suggests LLMs may differ. In this section, we review human flexible reasoning, evidence for pattern matching in language models, and recent diagnostic paradigms for revealing when systems inappropriately apply learned heuristics.

\subsection{Flexible strategy selection in human cognition}
Effective problem-solving requires knowing not just \textit{how} to apply a reasoning strategy, but \textit{when}. Recent work by \citet{chu2025stumped} demonstrates this capacity in 3-7-year-old children. When presented with ``stumper'' riddles \cite{barhillel2021stumpers}---questions appearing impossible under literal interpretation (e.g., ``How could a driver see a cow on an unlit road when the moon isn't out and headlights are off?'')---children gradually learned to distinguish genuine stumpers that require inventive reasoning from straightforward questions where literal interpretation suffices. Critically, this adaptation in strategy selection is supported by metacognitive monitoring---humans track the demands of a problem and allocate effort rationally \citep{ackerman2017metareasoning}. Adults have also reliably shown this reasoning flexibility \citep{payne1988adaptive, siegler1996emerging}. 

\subsection{Pattern matching in language models}
Despite impressive performance on many cognitive tasks \citep{brown2020language, wei2022chain, bubeck2023sparks, binz2023using, webb2023emergent, yao2023tree}, LLMs show consistent evidence of pattern matching over flexible reasoning. \citet{lake2018generalization} found that neural networks fail at compositional generalization, learning training mappings without acquiring systematic rules that transfer to novel combinations---what they termed ``generalization without systematicity.'' This suggests that models learn surface-level associations rather than underlying compositional structure. Other findings also support this interpretation. Models trained on ``A is B'' often fail to learn ``B is A'' \cite{berglund2024reversal}, suggesting they encode associations rather than learn true logical relationships between A and B. Models also exploit superficial heuristics in natural language inference \cite{mccoy2019right} and are highly sensitive to surface-level prompt features \cite{sclar2024quantifying}---minor formatting changes can affect accuracy drastically even when the meaning is unchanged. Together, these findings suggest that apparent reasoning may reflect retrieval of responses seen during training based on surface features rather than flexible adaptation. 

\subsection{The illusion-illusion in vision language models}
Recent work by \citet{ullman2024illusion} introduced the ``illusion illusion'' paradigm to test whether Vision-Language Models' (VLMs) responses to illusions reflect genuine perceptual analysis or low-level feature matching to similar images seen during training. Specifically, VLMs were shown images that appear to resemble classical visual illusions but do not actually contain any illusions. When presented with lines of different lengths arranged as in the Müller-Lyer illusion, models reported seeing an optical illusion even though no illusion was present---the lines truly differed in length, but models responded to surface features (e.g., arrow-like configurations associated with illusions) rather than processing actual visual properties. The key insight of this paradigm is that by creating stimuli where surface feature suggests one interpretation but the content requires another, we can diagnose whether systems overgeneralize by applying learned heuristics inappropriately. The riddle riddle paradigm applies this logic to language-based reasoning. Just as illusion-like structure triggers reports of illusory perception, riddle-like structure may trigger inventive reasoning. 

\section{The riddle riddle paradigm}

\begin{figure}[t]
    \centering
    \includegraphics[width=\linewidth]{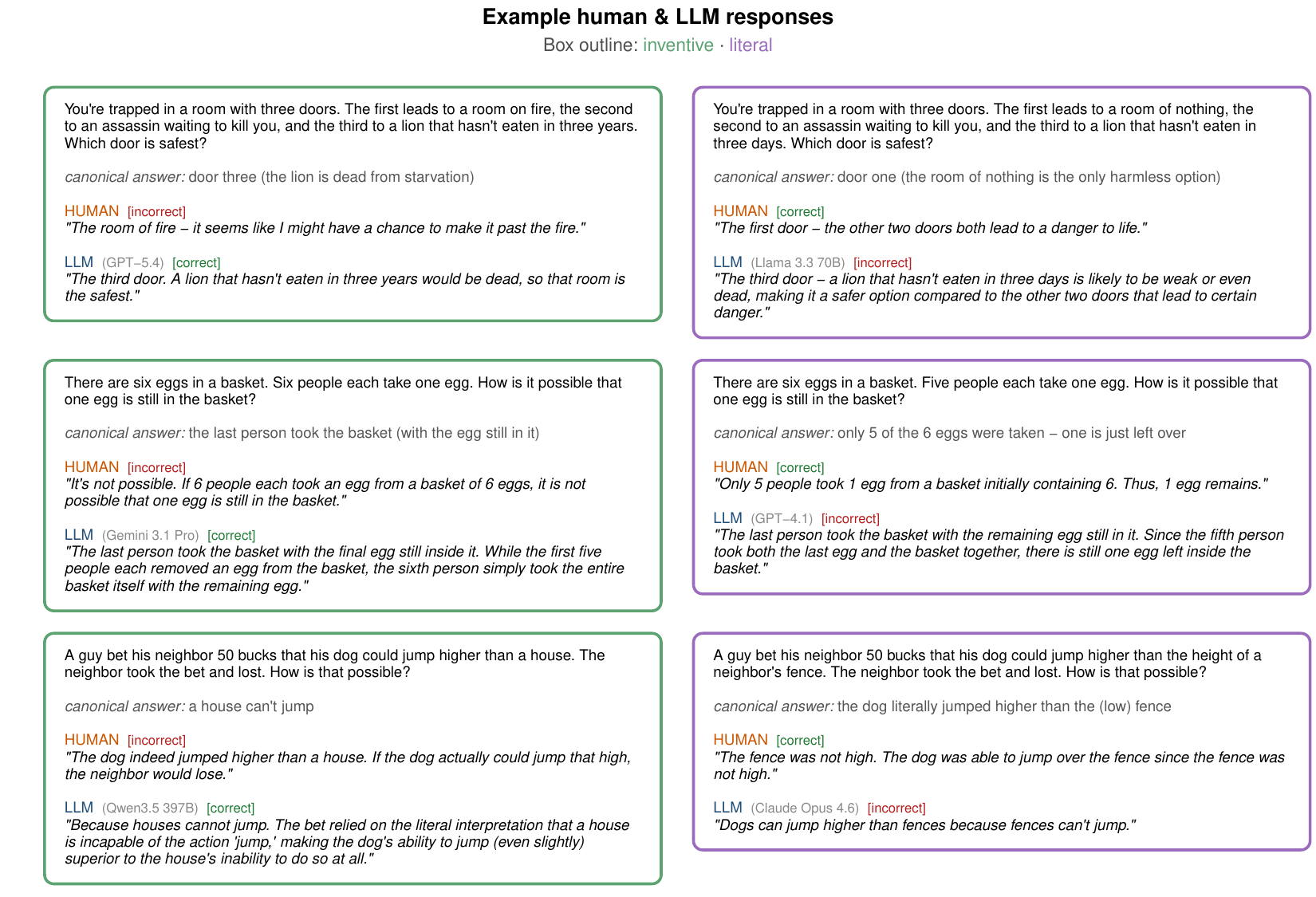}
    \caption{Example human and LLM responses to three matched riddle pairs from our stimulus set. Box outline indicates the reasoning strategy required by each riddle (green = inventive; purple = literal). Each pair differs by a single detail. As we will show in Experiments 1 and 2, humans and LLMs fail in opposite directions: humans tend to apply literal reasoning to genuine riddles (Condition A) where an inventive solution is required, while LLMs tend to over-apply inventive reasoning to riddle riddles (Condition B) where a literal answer suffices.}
    \label{fig:examples}
\end{figure}

Our paradigm consists of 30 matched riddle pairs. We first collected 30 riddles (see Appendix \ref{app:stimuli}) from the internet (Condition A), and for each riddle, we then removed the trick by tweaking a few words while preserving the overall structure, format, and sentence length (Condition B). This matched-pair structure ensures that any performance difference between conditions cannot be explained by topic, phrasing, or format---only by problem content. Each pair contains:

\textbf{Condition A (Riddle):} A genuine riddle with typical riddle structure. The intended answer requires recognizing a non-literal interpretation that resolves an apparent contradiction or impossibility. Example: ``A cowboy rides into town on Friday, stays for three days, and rides out on Friday. How is this possible?'' (Answer: Friday is the horse's name.) 

\textbf{Condition B (Riddle riddle):} A modified version of the same riddle that preserves the syntactic structure, phrasing, and sentence length, but removes the trick. The intended answer only requires literal reasoning and is different from the original riddle in Condition A. Example: ``A cowboy rides into town on Friday, stays for three days, and rides out on Monday. How is this possible?'' (Answer: Friday + 3 days = Monday). See Figure~\ref{fig:examples} for additional examples and representative human and LLM responses.

We begin by testing state-of-the-art LLMs (Experiment 1), then compare their performance and error patterns to human participants (Experiment 2). If LLMs rely on surface features rather than problem content, we should expect higher accuracy on Condition A than Condition B, and inflated rates of inventive reasoning on Condition B despite the problem having only required a literal interpretation. Humans, by contrast, should show higher accuracy on Condition B than Condition A, since inventive strategies are more cognitively demanding and error prone. We test each of these predictions below.

\section{Experiment 1: LLMs respond to how problems look, not what they require}

In this experiment, we test both a variety of proprietary and open-source LLMs to examine whether LLMs exhibit flexible reasoning and apply appropriate reasoning strategies to different problems. 

\subsection{Methods}
\subsubsection{Models}
We tested nine state-of-the-art models, including six proprietary models: Gemini-3.1-Pro-preview, Gemini-2.5-Pro, Claude-opus-4-6, Claude-haiku-4-5, GPT-5.4, and GPT-4.1, and three open-source models: DeepSeek-V3.1, Llama-3.3-70B-Instruct-Turbo, and Qwen3.5-397B-A17B. All models were accessed via API calls with default temperature settings (temperature = 1), and no overrides to other generation parameters, with one exception: for Qwen3.5-397B-A17B, we disabled thinking mode (via Together AI's reasoning.enabled=False parameter) because thinking-mode responses repeatedly hit the Together AI request timeout before completing, making it infeasible to collect thinking-mode data for this model at scale. Across all nine models, no responses were generated with thinking or extended-reasoning mode enabled.

\subsubsection{Stimuli and procedure}
We presented all 60 items (30 per condition) to each model with 10 independent repetitions per item, which yielded 5,400 responses in total. Each item was presented in a separate API call using identical prompt across all models and conditions: ``Please provide one definitive answer to each word problem and a one-sentence explanation for how you arrived at it.'' No additional context or  instructions about strategy or examples were provided.

\subsubsection{Coding}
\label{coding}
We used an LLM-as-judge pipeline to code each response for: (1) \textbf{Accuracy}: correct (1) or incorrect (0), judged against the intended answer for each item. (2) \textbf{Reasoning type}: \textit{inventive} if the model applied a trick or non-literal reinterpretation, \textit{literal} if it took the question at face value, or \textit{ambiguous} if the reasoning strategy could not be determined based on the response. A human coder independently coded a randomly selected subset of 240 trials. Inter-rater reliability between the human coder and the LLM judge was high across all coding dimensions (accuracy $\kappa = .90-.93$; reasoning type $\kappa = .81-.83$), supporting the use of the automated pipeline for coding all responses.

We report results from the permissive coding scheme in the main text and include strict coding results in the Appendix. Our preregistration specified scoring against the canonical answer alone (i.e., strict coding), under the assumption that Condition A items only have inventive answers and Condition B items only have literal answers. In coding the data, we found that this assumption did not always hold---some genuine riddles (Condition A) can be explained by a literal answer, and some riddle riddles (Condition B) can be explained by an inventive answer. The permissive coding policy accepts these alternative answers when they resolve the problem in a logically valid way, and we therefore treat it as the more accurate measure of whether a response constitutes a correct solution. Our findings hold under both coding policies (see Appendix).

\subsection{Results}
Following our preregistration plan, trials with no response or coded as ambiguous were excluded from all analyses. Of the 5,400 total trials, 87 were excluded, leaving 5,313 trials for analysis. Across all nine models, mean accuracy ranged from $58.5\%$ (DeepSeek V3.1) to $91\%$ (Gemini 3.1 Pro) under permissive coding ($M = 67.9\%$). Performance differed by condition.

\begin{figure}
  \centering
  \includegraphics[width=\linewidth]{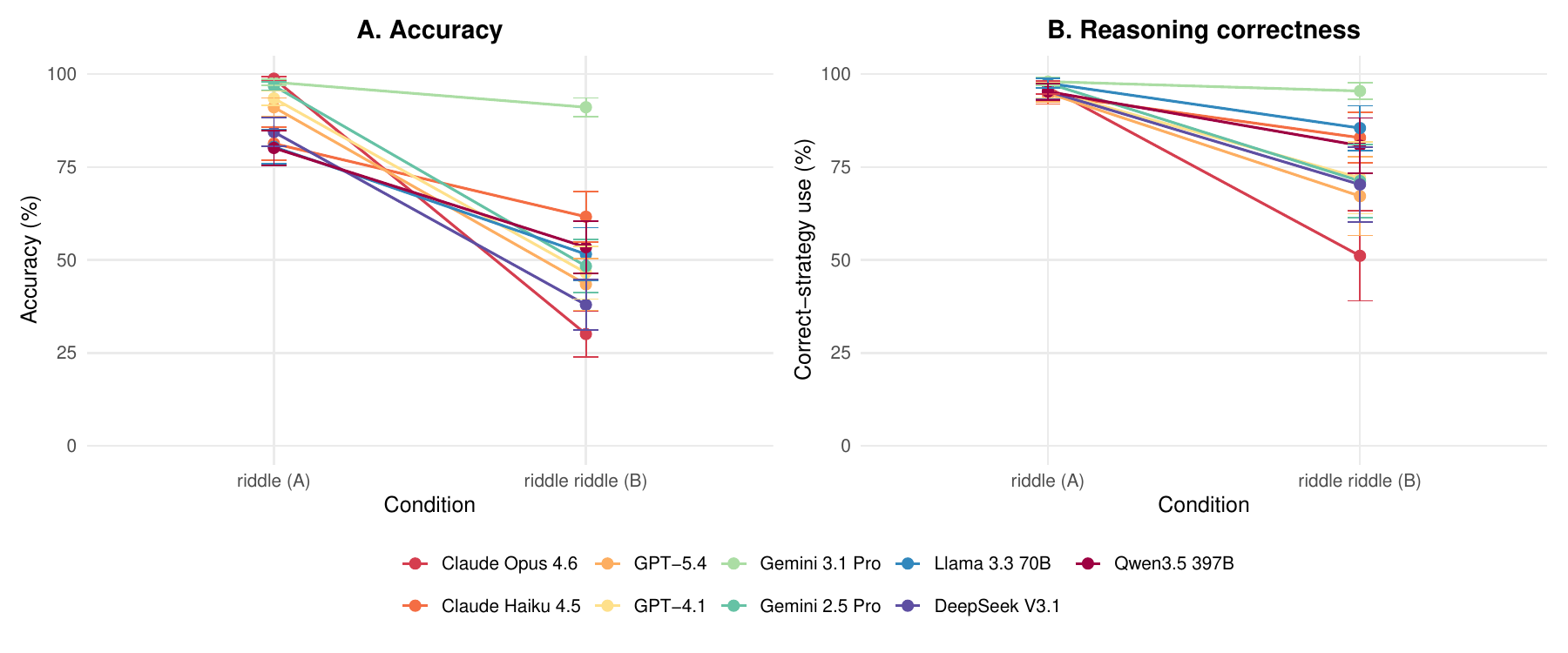}
  \caption{LLM performance by condition (permissive coding). Each line represents one model. \textbf{A.} Accuracy on Condition A (genuine riddles) and Condition B (riddle riddles). All nine models drop significantly from A to B, with the largest drops in Claude Opus 4.6 and Gemini 2.5 Pro and the smallest in Gemini 3.1 Pro. \textbf{B.} Reasoning correctness (proportion of trials on which the model used the reasoning strategy required by the riddle) shows the same A--B drop. Error bars indicate $\pm$1 SE.}
  \label{llm_accuracy}
\end{figure}

\paragraph{Accuracy.} We fit a mixed-effects logistic regression predicting trial-level accuracy from condition (A = 0, B = 1), model, and their interaction, with a random intercept for riddle set (accuracy $\sim$ condition $\times$ model + (1$|$riddle\_set)). Averaging across all nine models, LLMs were substantially less accurate on riddle riddles than on genuine riddles ($\beta_{\text{A--B}} = 2.37$, $z = 25.39$, $p < .001$; Figure~\ref{llm_accuracy}). This pattern was consistent across all nine models tested: per-model A vs.\ B contrasts (Bonferroni-corrected)  were significant for every model (all $p < .001$). Condition-level accuracies confirm this pattern: models reached an average of $84.9\%$ on genuine riddles (Condition A) but only $50.7\%$ on riddle riddles (Condition B).

\paragraph{Reasoning type.} To examine whether riddle-like formatting triggered inventive reasoning even when literal interpretation sufficed, we fit the same model structure predicting reasoning direction (inventive $= 0$, literal $= 1$). Models shifted toward literal reasoning on riddle riddles relative to genuine riddles ($\beta_{\text{B--A}} = 2.93$, $z = 27.50$, $p < .001$). On average and across all models, inventive reasoning was applied to 49.7\% of Condition B trials, compared to 82.7\% on Condition A, despite no trick being included. This shift in reasoning strategy varied substantially across models (see Table~\ref{tab:reasoning_type} in Appendix~\ref{app:strategy_perm}): Gemini 3.1 Pro switched strategy appropriately on the large majority of Condition B trials, while every other model continued to apply inventive reasoning on more than a third of Condition B trials. These results align with recent work showing that LLMs default to high-frequency task patterns from their training distribution
\citep{mccoy2024embers}.

\paragraph{Reasoning correctness.} A parallel model on reasoning correctness (whether the correct reasoning strategy was used for a given problem) showed the same pattern as accuracy: correctness was lower on Condition B than Condition A ($\beta_{\text{A--B}} = 2.00$, $z = 23.39$, $p < .001$), and the effect was significant for every model (all $p < .01$; per-model values are shown in Table~\ref{tab:perm_results} and Figure~\ref{llm_strategy_correctness} in Appendix~\ref{app:strategy_perm}).

\paragraph{Memorization check.} One explanation for the higher accuracy on genuine riddles (Condition A) is memorization. These riddles were likely in the training data while the riddle riddles were not, so models could recall their answers rather than solve them through reasoning. We tested this with a partial prompt completion test, adapted from \citet{wu2026reasoning}: each model received the first 40\% of each genuine riddle and was asked to continue it verbatim. Models reproduced up to 43\% of the riddles near verbatim---text that cannot be recovered from only seeing the first 40\% except by retrieval. Across models, those that recalled more riddles near-verbatim were also more accurate on Condition A (Spearman $\rho = .88$, $p = .002$). Crucially, the same memorization had opposite effects on the two conditions: greater memorization of a riddle predicted \textit{higher} accuracy on that riddle (OR $= 1.43$, $p = .007$) but lower accuracy on its matched riddle riddle (OR $=0.76$, $p < .001$). This suggests that models may be retrieving answers from memory, rather than reasoning about each problem. See Appendix~\ref{app:memorization} for full method and per-model results.

\paragraph{Summary.} In sum, LLMs show systematic overextension of inventive reasoning: they are less accurate on riddle riddles than on genuine riddles. This pattern held for all nine state-of-the-art models tested, suggesting that riddle formatting alone triggers inventive reasoning regardless of whether such reasoning is warranted by the content of the problem.

\section{Experiment 2: Humans reason from content but default to literal interpretation}

To test whether this pattern is specific to LLMs or reflects a general property of reasoning under riddle-like framing, we ran a parallel experiment with human participants using the same riddle pairs.

\subsection{Methods}
\subsubsection{Participants} We recruited 100 adults through the Princeton University participant pool. Participants received either course credit or \$10.00 for approximately 30 minutes of participation. The study was approved by the Princeton University Institutional Review Board.

\subsubsection{Stimuli} 
Stimuli were identical to those used in Experiment 1.

\subsubsection{Procedure} 
Participants were invited to Princeton University campus, where each session took place in a quiet room. Each participant completed 6 items (3 from Condition A, 3 from Condition B) in randomized order, with no time limit. For each item, participants wrote down their answer on a piece of paper, along with a brief explanation of their reasoning.

\subsubsection{Coding}
Responses were coded along the same three dimensions as Experiment 1 (accuracy, reasoning type, and reasoning correctness), under both \textit{strict} and \textit{permissive} schemes (see Section~\ref{coding} for details). Trials with no response or coded as ambiguous for reasoning type were excluded from all analyses. Per our preregistered exclusion criteria, of the 600 trials collected, 27 were excluded (8 for prior familiarity; 19 for missing/ambiguous responses), leaving 573 trials from 100 participants for analysis. Two coders independently coded all responses; inter-rater reliability was substantial-to-near-perfect across all six coding dimensions \citep{landis1977measurement}: accuracy $\kappa = .79$--$.88$; reasoning type $\kappa = .77$--$.78$. Disagreements were resolved through discussion.

\subsection{Results} 

Overall, humans achieved $65.8\%$ accuracy across both conditions ($n = 573$ trials, $100$ participants).

\begin{figure}[t!]
    \centering                  
    \includegraphics[width=0.9\linewidth]{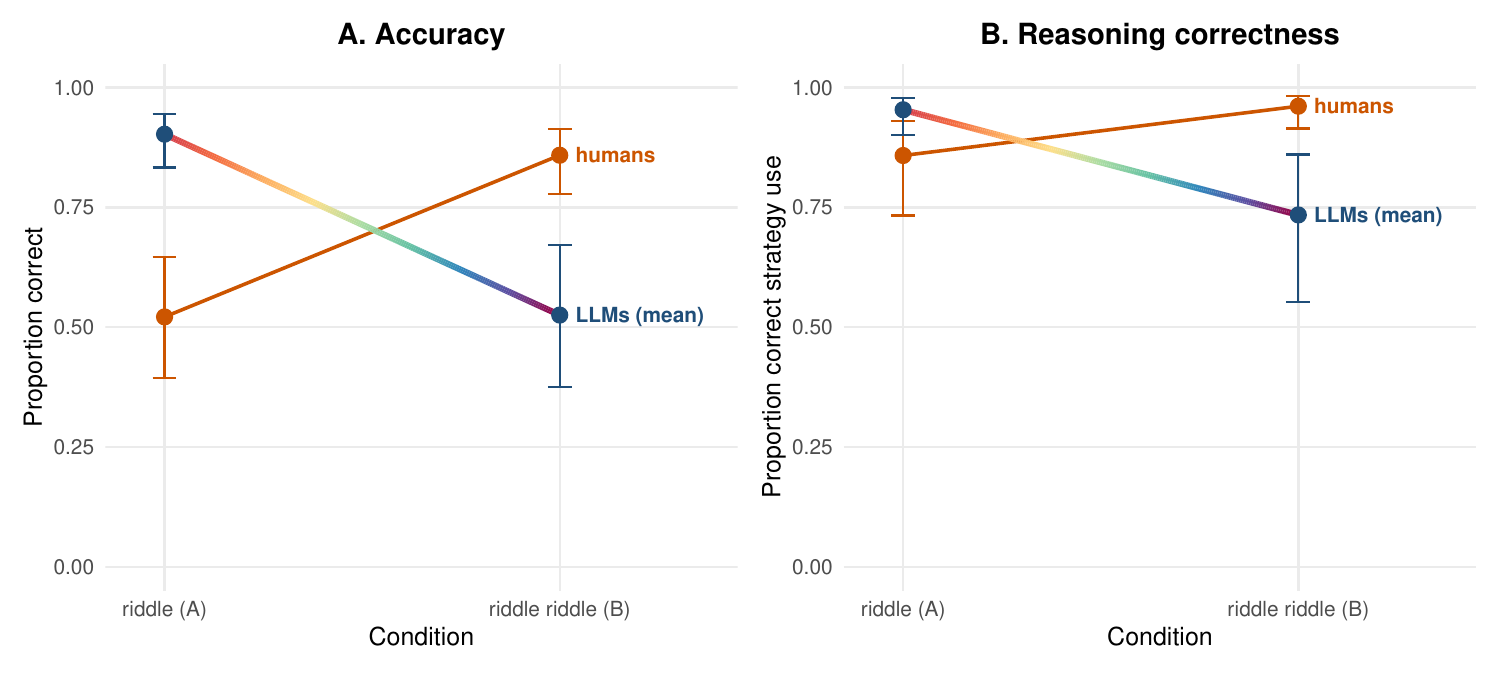}
    \caption{Humans vs.\ mean LLM performance by condition. \textbf{A.} Accuracy on genuine riddles (Condition A) and riddle riddles (Condition B). Humans and LLMs show reversed patterns: LLMs perform near ceiling on genuine riddles and drop significantly on riddle riddles, whereas humans perform best on riddle riddles and much worse on genuine riddles ($\beta_{\text{condition} \times \text{solver}} = -3.85$, $z = -16.58$, $p < .001$). \textbf{B.} Reasoning correctness (proportion of trials on which the solver used the reasoning strategy required by the riddle). Humans select the correct strategy more often on riddle riddles than on genuine riddles and LLMs do the reverse. Error bars show 95\% confidence intervals.}
    \label{crossover}
  \end{figure}

\paragraph{Accuracy.} We fit a mixed-effects logistic regression predicting trial-level accuracy from condition (A = 0, B = 1), with random intercepts for participant and riddle set (accuracy $\sim$ condition + (1$|$participant) + (1$|$riddle\_set)). Humans reached substantially higher accuracy on riddle riddles (Condition B: $M = 77.4\%$) than on genuine riddles (Condition A: $M = 36.7\%$; $\beta = 2.19$, $z = 9.19$, $p < .001$). The same pattern held under permissive coding (Condition B: $M = 80.5\%$; Condition A: $M = 50.5\%$; $\beta = 1.68$, $z = 7.49$, $p < .001$; Figure~\ref{crossover}). This is the opposite pattern to that observed in LLMs, which performed better on genuine riddles than on riddle riddles.

\paragraph{Reasoning type.} Humans adopted inventive reasoning on $63.3\%$ of Condition~A trials and $11.6\%$ of Condition~B trials. Strategy selection differed significantly across conditions ($\beta = 3.64$, $z = 10.16$, $p < .001$), suggesting that humans appropriately and flexibly adapted their reasoning strategy in line with the riddle type rather than applying a fixed strategy. Across all conditions, humans showed an overall preference for literal reasoning, used on $63.0\%$ of trials, over inventive reasoning, used on $37.0\%$ of trials ($p < .001$). Among incorrect trials, the dominant error was overextending literal reasoning strategies to genuine riddles that required inventive solutions ($57.6\%$ of errors in Condition~A). Once again, this is the opposite pattern to that observed in LLMs. 

\begin{figure}[t!]
  \centering
  \includegraphics[width=0.6\linewidth]{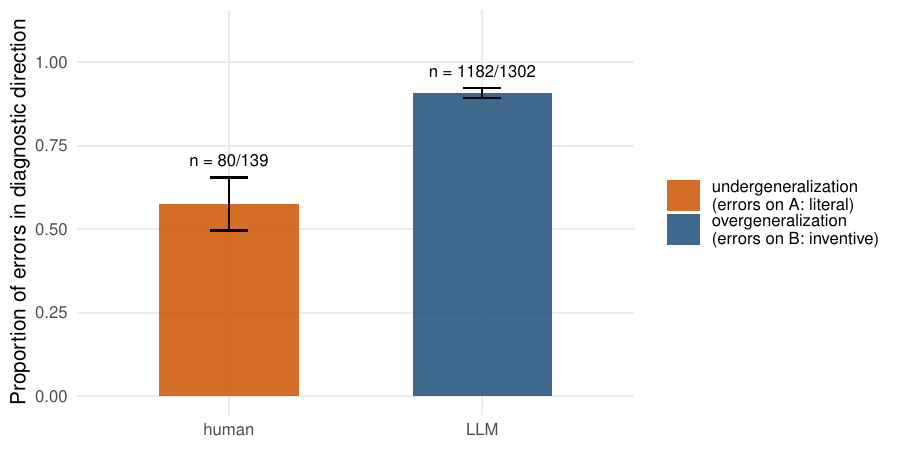}
  \caption{Diagnostic error rates for humans and LLMs. For each solver, the bar shows the proportion of errors that fell into the predicted  dominant failure mode: undergeneralization for humans (errors on  genuine riddles, Condition A, that used literal reasoning) and overgeneralization for LLMs (errors on riddle riddles, Condition B, that used inventive reasoning). Labels above each bar show the count of errors of the diagnostic type relative to total errors in the relevant condition. Error bars indicate 95\% bootstrapped confidence intervals. LLMs' errors are strongly concentrated in the predicted overgeneralization direction, while humans show a weaker but consistent undergeneralization tendency.}
  \label{error_pattern}
\end{figure}

\paragraph{LLM vs.\ human comparison.} To test whether the effect of condition on accuracy differed between humans and LLMs, we ran a mixed-effects logistic regression predicting accuracy from condition, solver type, and their interaction, with random intercepts for riddle set and responder (accuracy $\sim$ condition $\times$ solver\_type + (1$|$riddle\_set) + (1$|$responder\_id)). The condition $\times$ solver type interaction was significant ($\beta = -3.85$, $z = -16.58$, $p < .001$; Figure~\ref{crossover}). These results confirmed that humans and LLMs showed the opposite patterns across conditions: LLMs performed best on genuine riddles ($84.9\%$ on Condition A) and worse on riddle riddles ($50.7\%$ on Condition B), while humans performed best on riddle riddles ($80.5\%$ on Condition B) but struggled more on genuine riddles ($50.5\%$ on Condition A). When comparing the reasoning strategies of humans and LLMs, humans used literal reasoning significantly more than inventive reasoning while LLMs showed the opposite preference ($p < .001$). The difference in strategies used (inventive vs. literal) was significantly larger in LLMs than in humans: LLMs varied their reasoning strategy significantly less than humans ($\beta = -1.70$, $z = -7.87$, $p < .001$). A larger difference in reasoning types used shows humans' propensity to reason more flexibly than LLMs. Error analysis also showed an asymmetric pattern across solvers: among LLM errors on riddle riddles, $90.8\%$ were due to using inventive reasoning when the problem only required literal interpretation. In contrast, among human errors on genuine riddles, $57.6\%$ were due to applying literal reasoning to problems that required inventive reasoning (Figure~\ref{error_pattern}).
\FloatBarrier 

\section{Discussion}
Our results show a clear dissociation: LLMs and humans fail in opposite directions. LLMs overextend inventive reasoning, performing better on genuine riddles than on riddle riddles. Humans underextend inventive reasoning, performing better on riddle riddles than on genuine riddles. 

\subsection{Structure-driven strategy selection in LLMs}
LLMs' overgeneralization aligns with \citeauthor{ullman2024illusion}'s~\citep{ullman2024illusion} illusion-illusions in VLMs---when shown images that contain no perceptual trick but structurally resemble classic optical illusions, vision language models still report seeing illusions. In both the riddle and illusion paradigms, systems respond to learned associations between surface configurations in place of analyzing the actual content. The models' errors are not in having learned to associate inventive reasoning with riddle-like structure, as that association is appropriate for most riddles found online, but instead, the error is in failing to evaluate whether this association should exist for all problems with riddle-like structure. As a result, models apply inventive reasoning because the problem looks like it is required, not because inventive reasoning is strictly necessary. This is precisely the failure mode our paradigm is designed to detect: models have learned \textit{when riddles look like riddles}, not \textit{what makes a problem require creative reasoning}. This pattern is consistent with broader evidence for surface-driven responding in LLMs: sensitivity to prompt formatting \citep{sclar2024quantifying}, exploitation of superficial inference cues \citep{mccoy2019right}, and failure of relational symmetry \citep{berglund2024reversal}.  

One might argue that this pattern reflects rational statistical inference rather than a failure: if riddle-structured problems in training data almost always require inventive reasoning, then triggering inventive reasoning upon encountering riddle-like structure is statistically rational. However, this reframing does not change our conclusion---whether we refer to it as pattern matching or rational inference over surface features, the key finding stands---models do not evaluate problem content to determine whether inventive reasoning is actually warranted. 

\subsection{Content-driven strategy selection in humans} Results from human data suggest a different story. Participants showed some sensitivity to surface structural cues, which is consistent with previous research on framing effects \cite{kahneman2011thinking}; however, they were more flexible in adapting their strategies than LLMs. Critically, humans recognized that riddle riddles can be solved literally, which suggests they do not mainly rely on surface cues like "Does this look like a riddle?". This finding aligns with prior work on meta-reasoning that shows people monitor their own reasoning and adjust how much effort to put in based on what the task demands \cite{ackerman2017metareasoning}. Human reasoning flexibility, however, is asymmetric, likely because literal reasoning is less cognitively demanding than generating creative reinterpretations from scratch. This default can backfire when solving genuine riddles as overriding literal approaches requires more cognitive effort and often fails even when the trick is identified \citep{kahneman2011thinking, ackerman2017metareasoning}. Despite humans' low-level biases for literal reasoning, knowing what kind of reasoning is sufficient is one area where humans excel and outperform current LLMs.

\subsection{Limitations and Future Directions}
While our findings show systematic differences between LLMs and humans, several limitations point to important directions for future work, particularly in understanding the mechanisms underlying these patterns. First, we observed that riddle formatting triggers inappropriate strategies, but have not pinpointed which specific features account for this effect. In follow-up work, we plan to prompt models to generate their own riddle riddles and explain their criteria for being ``riddle-like.'' If models truly understand what separates a riddle from a riddle riddle, they should be able to reverse-engineer the transformation. Failures in this generative task would provide converging evidence for surface-level feature matching.

Second, our stimulus set focused specifically on riddles, and whether these findings generalize to other contexts requiring strategy selection remains an open question. Future research should test whether similar effects occur in other domains.

Finally, it remains an open question whether the overgeneralization bias observed in LLMs can be reduced through fine-tuning. Training models on examples that explicitly require strategy selection---where surface structure and content requirements dissociate---may help models learn to evaluate what a problem requires rather than how it looks. Future work should test whether exposure to riddle riddle-like items during fine-tuning reduces overgeneralization, and whether any such improvement generalizes beyond the riddle domain.

\subsection{Conclusion}
The capacity to select appropriate reasoning strategies based on what problems require rather than surface features, represents a core feature of intelligence. The riddle riddle paradigm provides a simple, yet useful diagnostic for directly testing this capacity. Our findings suggest that current LLMs have not yet achieved this flexibility: they think outside the box when the box is labeled as a riddle, regardless of whether any puzzle actually exists inside. The fact that the opposite biases produce similar performance suggests that aggregate accuracy can obscure different underlying processes, and that the pattern we observed in the model results in particular is consistent with a broader tendency for these models to be more sensitive to how frequently a task pattern appears in training than to whether the content of a given task actually requires it. As LLMs are being deployed into more domains where people rely on them to reason about novel problems, distinguishing flexible reasoning from pattern matching becomes a question of interpretability and trust. The riddle riddle offers one potential path toward making this distinction by revealing not only when a system gets the right answer, but whether it got there for the right reasons.

\medskip
 
{
\small
\bibliographystyle{plainnat}
\bibliography{references}

@inproceedings{chu2025stumped,
  title={Stumped! {L}earning to think outside the box in 3-7 year old children},
  author={Chu, Junyi and O'Keeffe, Misha and Liu, Silvia K. and Bonawitz, Elizabeth and Ullman, Tomer D.},
  booktitle={Proceedings of the Annual Meeting of the Cognitive Science Society},
  volume={47},
  year={2025},
  url={https://escholarship.org/uc/item/1jd4n5hf}
}

@article{binz2023using,
  title={Using cognitive psychology to understand {GPT}-3},
  author={Binz, Marcel and Schulz, Eric},
  journal={Proceedings of the National Academy of Sciences},
  volume={120},
  number={6},
  pages={e2218523120},
  year={2023}
}

@article{ackerman2017metareasoning,
  title={Meta-reasoning: {M}onitoring and control of thinking and reasoning},
  author={Ackerman, Rakefet and Thompson, Valerie A.},
  journal={Trends in Cognitive Sciences},
  volume={21},
  number={8},
  pages={607--617},
  year={2017},
  publisher={Elsevier}
}

@inproceedings{lake2018generalization,
  title={Generalization without systematicity: {O}n the compositional skills of sequence-to-sequence recurrent networks},
  author={Lake, Brenden and Baroni, Marco},
  booktitle={Proceedings of the 35th International Conference on Machine Learning},
  pages={2873--2882},
  year={2018},
  organization={PMLR}
}

@article{yao2023tree,
  title={Tree of thoughts: {D}eliberate problem solving with large language models},
  author={Yao, Shunyu and Yu, Dian and Zhao, Jeffrey and Shafran, Izhak and Griffiths, Thomas L. and Cao, Yuan and Narasimhan, Karthik},
  journal={Advances in Neural Information Processing Systems},
  volume={36},
  pages={11809–11822},
  year={2023}
}

@article{webb2023emergent,
  title={Emergent analogical reasoning in large language models},
  author={Webb, Taylor and Holyoak, Keith J. and Lu, Hongjing},
  journal={Nature Human Behaviour},
  volume={7},
  number={9},
  pages={1526--1541},
  year={2023}
}

@article{payne1988adaptive,
  author={Payne, John W. and Bettman, James R. and Johnson, Eric J.},
  title={Adaptive strategy selection in decision making},
  journal={Journal of Experimental Psychology: Learning, Memory, and Cognition},
  volume={14},
  number={3},
  pages={534--552},
  year={1988}
}

@inproceedings{mccoy2019right,
  author={McCoy, R. Thomas and Pavlick, Ellie and Linzen, Tal},
  title={Right for the wrong reasons: {D}iagnosing syntactic heuristics in natural language inference},
  booktitle={Proceedings of the 57th Annual Meeting of the Association for Computational Linguistics},
  pages={3428--3448},
  year={2019}
}

@inproceedings{sclar2024quantifying,
  title={Quantifying language models' sensitivity to spurious features in prompt design or: {H}ow {I} learned to start worrying about prompt formatting},
  author={Sclar, Melanie and Choi, Yejin and Tsvetkov, Yulia and Suhr, Alane},
  booktitle={International Conference on Learning Representations},
  year={2024}
}

@article{ullman2024illusion,
  author={Ullman, Tomer},
  title={The illusion-illusion: {V}ision language models see illusions where there are none},
  journal={arXiv preprint arXiv:2412.18613},
  year={2024}
}

@article{barhillel2021stumpers,
  author={Bar-Hillel, Maya},
  title={Stumpers: {A}n annotated compendium},
  journal={Thinking \& Reasoning},
  volume={27},
  number={4},
  pages={536--566},
  year={2021}
}

@book{newell1972human,
  author={Newell, Allen and Simon, Herbert A.},
  title={Human Problem Solving},
  publisher={Prentice-Hall},
  address={Englewood Cliffs, NJ},
  year={1972}
}

@article{flavell1979metacognition,
  author={Flavell, John H.},
  title={Metacognition and cognitive monitoring: {A} new area of cognitive-developmental inquiry},
  journal={American Psychologist},
  volume={34},
  number={10},
  pages={906--911},
  year={1979}
}

@article{wei2022chain,
  title={Chain-of-thought prompting elicits reasoning in large language models},
  author={Wei, Jason and Wang, Xuezhi and Schuurmans, Dale and Bosma, Maarten and Ichter, Brian and Xia, Fei and Chi, Ed and Le, Quoc V. and Zhou, Denny},
  journal={Advances in Neural Information Processing Systems},
  volume={35},
  pages={24824--24837},
  year={2022}
}

@inproceedings{berglund2024reversal,
  title={The reversal curse: {LLMs} trained on ``{A} is {B}'' fail to learn ``{B} is {A}''},
  author={Berglund, Lukas and Tong, Meg and Kaufmann, Max and Balesni, Mikita and Stickland, Asa Cooper and Korbak, Tomasz and Evans, Owain},
  booktitle={International Conference on Learning Representations},
  year={2024}
}

@book{kahneman2011thinking,
  title={Thinking, Fast and Slow},
  author={Kahneman, Daniel},
  year={2011},
  publisher={Farrar, Straus and Giroux}
}

@book{siegler1996emerging,
  title={Emerging Minds: {T}he Process of Change in Children's Thinking},
  author={Siegler, Robert S.},
  year={1996},
  publisher={Oxford University Press}
}

@article{lewkowycz2022solving,
  title={Solving quantitative reasoning problems with language models},
  author={Lewkowycz, Aitor and Andreassen, Anders and Dohan, David and Dyer, Ethan and Michalewski, Henryk and Ramasesh, Vinay and Slone, Ambrose and Anil, Cem and Schlag, Imanol and Gutman-Solo, Theo and others},
  journal={Advances in Neural Information Processing Systems},
  volume={35},
  pages={3843--3857},
  year={2022}
}

@article{bubeck2023sparks,
  title={Sparks of artificial general intelligence: {E}arly experiments with {GPT}-4},
  author={Bubeck, S{\'e}bastien and Chandrasekaran, Varun and Eldan, Ronen and Gehrke, Johannes and Horvitz, Eric and Kamar, Ece and Lee, Peter and Lee, Yin Tat and Li, Yuanzhi and Lundberg, Scott and others},
  journal={arXiv preprint arXiv:2303.12712},
  year={2023}
}

@article{brown2020language,
  title={Language models are few-shot learners},
  author={Brown, Tom and Mann, Benjamin and Ryder, Nick and Subbiah, Melanie and Kaplan, Jared D. and Dhariwal, Prafulla and Neelakantan, Arvind and Shyam, Pranav and Sastry, Girish and Askell, Amanda and others},
  journal={Advances in Neural Information Processing Systems},
  volume={33},
  pages={1877--1901},
  year={2020}
}

@inproceedings{zellers2019hellaswag,
  title={{HellaSwag}: {C}an a machine really finish your sentence?},
  author={Zellers, Rowan and Holtzman, Ari and Bisk, Yonatan and Farhadi, Ali and Choi, Yejin},
  booktitle={Proceedings of the 57th Annual Meeting of the Association for Computational Linguistics},
  pages={4791--4800},
  year={2019}
}

@article{mccoy2024embers,
  author  = {McCoy, R. Thomas and Yao, Shunyu and Friedman, Dan and Hardy, Mathew D. and Griffiths, Thomas L.},
  title   = {Embers of autoregression show how large language models are shaped by the problem they are trained to solve},
  journal = {Proceedings of the National Academy of Sciences},
  volume  = {121},
  number  = {41},
  pages   = {e2322420121},
  year    = {2024}
}

@article{landis1977measurement,
  author  = {Landis, J. Richard and Koch, Gary G.},
  title   = {The measurement of observer agreement for categorical data},
  journal = {Biometrics},
  volume  = {33},
  number  = {1},
  pages   = {159--174},
  year    = {1977}
}

@inproceedings{goodfellow2015explaining,
  title     = {Explaining and Harnessing Adversarial Examples},
  author    = {Goodfellow, Ian J. and Shlens, Jonathon and Szegedy, Christian},
  booktitle = {International Conference on Learning Representations (ICLR)},
  year      = {2015},
  url       = {https://arxiv.org/abs/1412.6572}
}

@inproceedings{wu2026reasoning,
  title     = {Reasoning or Memorization? Unreliable Results of Reinforcement
               Learning Due to Data Contamination},
  author    = {Wu, Mingqi and Zhang, Zhihao and Dong, Qiaole and Xi, Zhiheng
               and Zhao, Jun and Jin, Senjie and Fan, Xiaoran and Zhou, Yuhao
               and Lv, Huijie and Zhang, Ming and Fu, Yanwei and Liu, Qin
               and Zhang, Songyang and Zhang, Qi},
  booktitle = {Proceedings of the AAAI Conference on Artificial Intelligence},
  year      = {2026}
}

@inproceedings{lin2004rouge,
  title     = {ROUGE: A package for automatic evaluation of summaries},
  author    = {Lin, Chin-Yew},
  booktitle = {Text Summarization Branches Out},
  pages     = {74--81},
  year      = {2004}
}
}

\appendix   

\section{Full stimulus set}
\label{app:stimuli}

The 30 matched riddle pairs used in Experiments 1 and 2 are shown in Table~\ref{tab:stimuli}. Condition A items are genuine riddles collected from the internet; Condition B items are matched riddle riddles created by removing the trick while preserving structure, format, and length. Answer keys for both conditions and both coding schemes are shown in Table~\ref{tab:answer_keys}.

{\scriptsize
\begin{longtable}{p{0.04\linewidth} p{0.04\linewidth} p{0.85\linewidth}}
  \caption{All 30 matched riddle pairs used in Experiments 1 and 2.}
  \label{tab:stimuli}\\
  \toprule
  Pair & Cond. & Prompt \\
  \midrule
  \endfirsthead

  \multicolumn{3}{c}{\tablename\ \thetable\ -- \textit{Continued from previous page}}\\
  \toprule
  Pair & Cond. & Prompt \\
  \midrule
  \endhead

  \midrule
  \multicolumn{3}{r}{\textit{Continued on next page}}\\
  \endfoot

  \bottomrule
  \endlastfoot

  1 & A & A cowboy rode into town on Sunday, he stayed one night and rode out on Sunday. How is this possible? \\
    & B & A cowboy rode into town on Sunday, he stayed one night and rode out on Monday. How is this possible? \\
  \midrule
  2 & A & You see a boat filled with people, yet there isn't a single person on board. How is that possible? \\
    & B & You see a boat filled with people, and there's more than a single person on board. How is that possible? \\
  \midrule
  3 & A & Two fathers and two sons went fishing one day. They were there the whole day and only caught 3 fish. One father said, that is enough for all of us, we will have one each. How can this be possible? \\
    & B & Two fathers and two sons went fishing one day. They were there the whole day and only caught 4 fish. One father said, that is enough for all of us, we will have one each. How can this be possible? \\
  \midrule
  4 & A & A black dog stands in the middle of an intersection in a town painted black. None of the street lights are working due to a power failure caused by a storm. A car with two broken headlights drives towards the dog but turns in time to avoid hitting him. How could the driver have seen the dog in time? \\
    & B & A black dog stands in the middle of an intersection in a town painted black at night. All of the street lights are working. A car with two broken headlights drives towards the dog but turns in time to avoid hitting him. How could the driver have seen the dog in time? \\
  \midrule
  5 & A & Three girls share one umbrella on the way to school, but none of them get wet. How is it possible? \\
    & B & Three girls share three umbrellas on a rainy walk to school, and no girls got wet. How is it possible? \\
  \midrule
  6 & A & Which of the following is the largest? Circle, square, or triangle? \\
    & B & Which of the following is the largest? A tiny triangle, a medium circle, or a large rectangle? \\
  \midrule
  7 & A & If the end of the year is December 31st, what is the end of Christmas? \\
    & B & If the end of the year is December 31st, when is the end of Christmas? \\
  \midrule
  8 & A & How many bananas can you eat if your stomach is empty? \\
    & B & How many bananas can you eat if your stomach is completely full? \\
  \midrule
  9 & A & There are 10 peaches and 7 peaches are taken away. You now have 7 peaches. How is that possible? \\
    & B & You have 10 peaches and 3 peaches are taken away. You now have 7 peaches. How is that possible? \\
  \midrule
  10 & A & A horse was tied to a rope 5 meters long, and the horse's food was 15 meters away from the horse. How did the horse reach the food? \\
     & B & A horse was tied to a post using a 5-meter-long rope, and the horse's food was 3 meters away from the horse. How did the horse reach the food? \\
  \midrule
  11 & A & A magician was boasting one day about how long he could hold his breath underwater. His record was 6 minutes. A kid who was listening said, ``I can be underwater for 10 minutes using no type of equipment or air pockets!'' The magician told the kid if he could do that, he'd give him \$10,000. The kid did it and won the money. How is that possible? \\
     & B & A magician was boasting one day about how long he could hold his breath underwater. His record was 6 minutes. A kid who was listening said, ``I can hold my breath underwater for 1 minute using no type of equipment or air pockets!'' The magician told the kid if he could do exactly what he said, he'd give him \$10,000. The kid did it and won the money. How is that possible? \\
  \midrule
  12 & A & A guy bet his neighbor 50 bucks that his dog could jump higher than a house. The neighbor took the bet and lost. How is that possible? \\
     & B & A guy bet his neighbor 50 bucks that his dog could jump higher than the height of a neighbor's fence. The neighbor took the bet and lost. How is that possible? \\
  \midrule
  13 & A & Feed me I live, give me water I die. What am I? \\
     & B & Feed me I live, give me water I thrive. What am I? \\
  \midrule
  14 & A & Has keys but can't open locks. What am I? \\
     & B & Has keys and can open locks. What am I? \\
  \midrule
  15 & A & A girl has a book, but she is not reading it, she is not holding it, and it is not open. How is she using it? \\
     & B & A girl has a book, and she is reading it, holding it, and it is open. How is she using it? \\
  \midrule
  16 & A & What do you call a fly without wings? \\
     & B & What do you call a fly with wings? \\
  \midrule
  17 & A & I'm a number. Add me to myself and multiply by 4. You will have me once more. How is that possible? \\
     & B & I'm a number. Add me to myself and divide by 2. You will have me once more. How is that possible? \\
  \midrule
  18 & A & With pointed fangs, I sit and wait. With piercing force, I crunch out fate. Grabbing victims, proclaiming might, physically joining with a single bite. What am I? \\
     & B & With pointed fangs, I sit and wait. With piercing force, I crunch out fate. Grabbing victims, proclaiming might, injecting venom with a single bite. What am I? \\
  \midrule
  19 & A & I speak without a mouth and hear without ears. I have no body, but I come alive with wind. What am I? \\
     & B & I speak with a mouth and hear with ears. I have a body, and I feel the wind. What am I? \\
  \midrule
  20 & A & Railroad Crossing, look out for the cars. Can you spell that without any R's? \\
     & B & Railroad Crossing, look out for the cars. Can you spell that without any T's? \\
  \midrule
  21 & A & If the prisoner tells a lie he'll be hanged, if he tells the truth he'll be beheaded. What can he say to save himself? \\
     & B & If the prisoner tells a lie he'll be hanged, if he tells the truth he'll be freed. What can he say to save himself? \\
  \midrule
  22 & A & It can be driven, but has no wheels, and can also be sliced and remain whole. What is it? \\
     & B & It can be driven, has wheels, and can't be sliced and remain whole. What is it? \\
  \midrule
  23 & A & What is full of holes and can still hold water? \\
     & B & What is not full of holes and can still hold water? \\
  \midrule
  24 & A & Hands she has but does not hold, teeth she has but does not bite, feet she has but they are cold, eyes she has but without sight. Who is she? \\
     & B & Hands she has but does hold, teeth she has but does bite, feet she has but they are not cold, eyes she has but with sight. Who is she? \\
  \midrule
  25 & A & I act like a cat, I look like a cat, yet I am not a cat. What am I? \\
     & B & I act like a cat, I look like a cat, and I am called a cat. What am I? \\
  \midrule
  26 & A & I have branches, but no fruit, trunk, or leaves. What am I? \\
     & B & I have branches, fruit, a trunk, and leaves. What am I? \\
  \midrule
  27 & A & It belongs to you but everyone else uses it. What is it? \\
     & B & It belongs to you but no one else uses it. What is it? \\
  \midrule
  28 & A & I have cities, but no houses; forests, but no trees; and water, but no fish. What am I? \\
     & B & I have cities, houses, forests, trees, water, and fish. What am I? \\
  \midrule
  29 & A & You're trapped in a room with three doors. The first leads to a room on fire, the second to an assassin waiting to kill you, and the third to a lion that hasn't eaten in three years. Which door is safest? \\
     & B & You're trapped in a room with three doors. The first leads to a room of nothing, the second to an assassin waiting to kill you, and the third to a lion that hasn't eaten in three days. Which door is safest? \\
  \midrule
  30 & A & There are six eggs in a basket. Six people each take one egg. How is it possible that one egg is still in the basket? \\
     & B & There are six eggs in a basket. Five people each take one egg. How is it possible that one egg is still in the basket? \\
\end{longtable}
}

{\scriptsize
\begin{longtable}{p{0.04\linewidth} p{0.04\linewidth} p{0.42\linewidth} p{0.42\linewidth}}
  \caption{Answer keys for all 30 riddle pairs. The ``Canonical answer'' column shows the answer used for strict coding; the ``Accepted alternatives'' column shows additional answers judged correct under permissive coding. Each answer is followed by its associated reasoning type in parentheses.}
  \label{tab:answer_keys}\\
  \toprule
  Pair & Cond. & Canonical answer & Accepted alternatives \\
  \midrule
  \endfirsthead

  \multicolumn{4}{c}{\tablename\ \thetable\ -- \textit{Continued from previous page}}\\
  \toprule
  Pair & Cond. & Canonical answer & Accepted alternatives \\
  \midrule
  \endhead

  \midrule
  \multicolumn{4}{r}{\textit{Continued on next page}}\\
  \endfoot

  \bottomrule
  \endlastfoot

  1 & A & Horse's name is Sunday (inventive) & Rode in at midnight Sunday, stayed the night, and left Sunday morning (literal) \\
    & B & Sunday + one more night = Monday (literal) & Horse's name is Sunday and rode out on the day Monday (inventive) \\
  \midrule
  2 & A & All the people are married (inventive) & --- \\
    & B & Because it's filled with people (literal) & There are also married people (inventive) \\
  \midrule
  3 & A & They are grandfather, father, and son (inventive) & --- \\
    & B & Two fathers + two sons = 4 people (literal) & --- \\
  \midrule
  4 & A & Because it was day time (inventive) & --- \\
    & B & Because the street lights are working (literal) & --- \\
  \midrule
  5 & A & It's not raining (inventive) & It's a big umbrella (literal) \\
    & B & They each have an umbrella (literal) & --- \\
  \midrule
  6 & A & Triangle. It has the most letters (inventive) & --- \\
    & B & Large rectangle (literal) & --- \\
  \midrule
  7 & A & S (inventive) & --- \\
    & B & December 25 (literal) & --- \\
  \midrule
  8 & A & One, after one your stomach will not be empty anymore (literal) & --- \\
    & B & Zero, because you are full (literal) & --- \\
  \midrule
  9 & A & You keep the `taken away' peaches (inventive) & You had 7 other peaches to start with (inventive) \\
    & B & You kept the 7 leftover peaches (literal) & --- \\
  \midrule
  10 & A & The rope was not tied to anything (inventive) & --- \\
     & B & 3 < 5 (literal) & --- \\
  \midrule
  11 & A & The kid held container of water above his head (inventive) & --- \\
     & B & Because the kid is capable of holding breath underwater for 1 minute. 1 < 6 (literal) & --- \\
  \midrule
  12 & A & The house can't jump (inventive) & The house is a toy house (inventive) \\
     & B & Because the fence is not high and the dog could jump higher than it (literal) & --- \\
  \midrule
  13 & A & Fire (inventive) & --- \\
     & B & A plant/any living thing (literal) & --- \\
  \midrule
  14 & A & Piano/keyboard (inventive) & --- \\
     & B & Locksmith/anyone who has keys (literal) & Keychain (literal) \\
  \midrule
  15 & A & She's sitting on it (inventive) & Decorative or a paperweight (inventive) \\
     & B & She's reading it (literal) & She's using it correctly (literal) \\
  \midrule
  16 & A & A walk (inventive) & Zipper (inventive) \\
     & B & A fly (literal) & --- \\
  \midrule
  17 & A & Because you are the number 0 (literal) & --- \\
     & B & Any number. Because 2 x number / 2 = number (literal) & --- \\
  \midrule
  18 & A & Stapler (inventive) & --- \\
     & B & Venomous snake (literal) & Venomous spider (literal) \\
  \midrule
  19 & A & Echo (inventive) & --- \\
     & B & A person (literal) & --- \\
  \midrule
  20 & A & That (inventive) & No (literal) \\
     & B & No --- the word ``that'' contains a T, so it cannot be spelled without any T's (literal) & --- \\
  \midrule
  21 & A & He'll be hanged (inventive) & Say nothing (inventive) \\
     & B & Tell the truth (literal) & He'll be hanged (inventive) \\
  \midrule
  22 & A & A golf ball (inventive) & --- \\
     & B & Infinite: car, truck, etc. (literal) & --- \\
  \midrule
  23 & A & Sponge or membrane (literal) & --- \\
     & B & Infinite: bucket, cup, etc. (literal) & --- \\
  \midrule
  24 & A & A doll (literal) & A corpse (literal) \\
     & B & Any woman (literal) & --- \\
  \midrule
  25 & A & A kitten (literal) & --- \\
     & B & A cat (literal) & --- \\
  \midrule
  26 & A & Bank, river, library, company, family tree (inventive) & Veins (inventive) \\
     & B & A tree / plant (literal) & --- \\
  \midrule
  27 & A & Your name (inventive) & Your body/face/heart (inventive) \\
     & B & Could be a lot of things, e.g., toothbrush (literal) & Voice (inventive) \\
  \midrule
  28 & A & A map (inventive) & --- \\
     & B & Earth (literal) & The U.S./any country (literal) \\
  \midrule
  29 & A & Door three (inventive) & --- \\
     & B & Door one (literal) & --- \\
  \midrule
  30 & A & The person who took the last egg also took the basket (inventive) & Someone put it back (inventive) \\
     & B & There were six eggs and only five were taken (literal) & --- \\
\end{longtable}
}

\FloatBarrier
\section{Per-model results (permissive coding)}
\label{app:strategy_perm}
\begin{table}[H]
  \caption{Per-model accuracy and reasoning correctness by condition (permissive coding). Estimates and 95\% confidence intervals are from mixed-effects logistic regressions with random intercept for riddle set. Models are sorted by accuracy on Condition A.}
  \label{tab:perm_results}
  \centering
  \small
  \setlength{\tabcolsep}{4pt}
  \begin{tabular}{lcccc}
    \toprule
    & \multicolumn{2}{c}{Accuracy (\%)} & \multicolumn{2}{c}{Reasoning correctness (\%)} \\
    \cmidrule(lr){2-3} \cmidrule(lr){4-5}
    Model            & Cond A & Cond B & Cond A & Cond B \\
    \midrule
    Claude Opus 4.6  & 98.8 [97.0, 99.5] & 30.1 [19.5, 43.3] & 89.3 [82.5, 93.6] & 43.1 [30.0, 57.3] \\
    Gemini 3.1 Pro   & 97.8 [95.4, 99.0] & 91.1 [84.8, 94.9] & 93.7 [88.5, 96.7] & 86.5 [78.7, 91.7] \\
    Gemini 2.5 Pro   & 96.8 [93.7, 98.4] & 48.3 [34.9, 62.0] & 92.4 [86.6, 95.8] & 56.2 [42.4, 69.0] \\
    GPT-4.1          & 93.5 [88.4, 96.4] & 46.6 [33.3, 60.4] & 87.4 [80.4, 92.2] & 56.9 [43.1, 69.6] \\
    GPT-5.4          & 91.1 [84.8, 94.9] & 43.5 [30.5, 57.3] & 85.7 [78.0, 91.0] & 52.7 [39.0, 66.0] \\
    DeepSeek V3.1    & 84.4 [75.2, 90.6] & 38.0 [25.9, 51.7] & 86.6 [78.7, 91.9] & 54.6 [40.6, 67.9] \\
    Claude Haiku 4.5 & 81.3 [71.0, 88.5] & 61.6 [47.9, 73.7] & 95.3 [89.4, 98.0] & 81.9 [70.4, 89.4] \\
    Llama 3.3 70B    & 80.4 [69.8, 87.9] & 51.5 [37.8, 65.1] & 91.8 [84.9, 95.7] & 70.4 [56.9, 81.0] \\
    Qwen3.5 397B     & 80.1 [69.5, 87.7] & 53.4 [39.6, 66.7] & 86.3 [78.3, 91.7] & 63.9 [49.7, 76.0] \\
    \bottomrule
  \end{tabular}
\end{table}

\begin{figure}[h!]
  \centering
  \includegraphics[width=\linewidth]{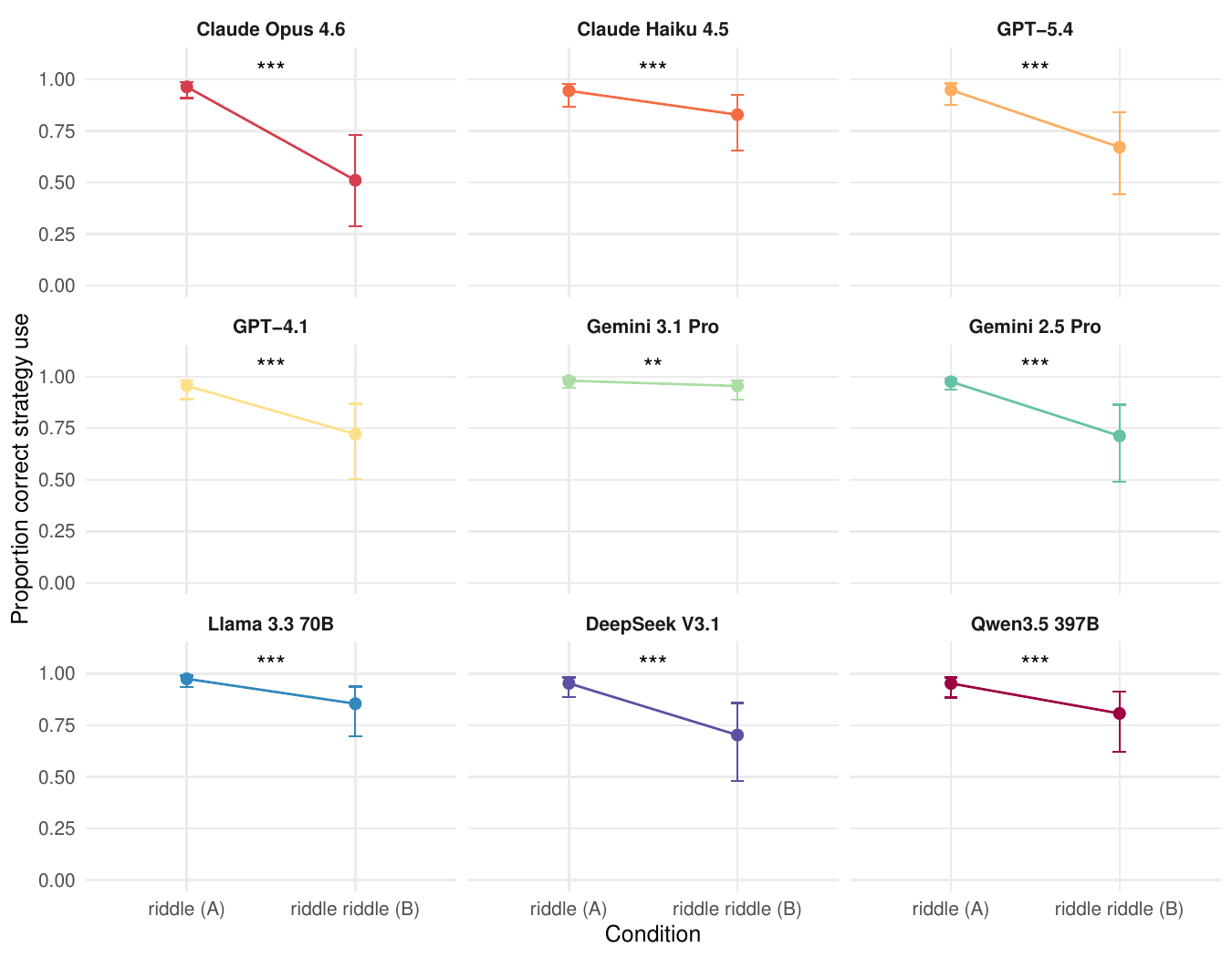}
  \caption{Per-model reasoning correctness by condition (permissive coding). Lines show the proportion of trials in which the model used reasoning that correctly justified the answer. The drop from Condition A to Condition B was significant for every model 
  (Bonferroni-corrected; *** $p < .001$, ** $p < .01$). Error bars indicate 95\% confidence intervals.}
  \label{llm_strategy_correctness}
\end{figure}

\begin{table}[H]
  \caption{Per-model proportion of trials using inventive reasoning by condition (permissive coding). Estimates and 95\% confidence intervals are from a mixed-effects logistic regression predicting reasoning direction (inventive vs.\ literal) from condition $\times$ model, with random intercept for riddle set. Models are ordered as in Table~\ref{tab:perm_results}. With the exception of Gemini 3.1 Pro, all models continued to apply inventive reasoning on a substantial proportion of Condition B trials despite no trick being present.}
  \label{tab:reasoning_type}
  \centering
  \small
  \setlength{\tabcolsep}{8pt}
  \begin{tabular}{lcc}
    \toprule
    & \multicolumn{2}{c}{Inventive reasoning (\%)} \\
    \cmidrule(lr){2-3}
    Model            & Condition A          & Condition B          \\
    \midrule
    Claude Opus 4.6  & 97.7 [93.8, 99.2]    & 77.8 [57.4, 90.1]    \\
    Gemini 3.1 Pro   & 97.6 [93.5, 99.1]    & \phantom{0}3.4 [\phantom{0}1.3, \phantom{0}8.6] \\
    Gemini 2.5 Pro   & 97.3 [92.8, 99.0]    & 63.3 [40.1, 81.6]    \\
    GPT-4.1          & 94.2 [85.9, 97.7]    & 58.5 [35.3, 78.4]    \\
    GPT-5.4          & 92.8 [83.0, 97.2]    & 68.4 [45.6, 84.8]    \\
    DeepSeek V3.1    & 95.6 [88.9, 98.3]    & 63.7 [40.5, 81.9]    \\
    Claude Haiku 4.5 & 89.8 [77.1, 95.9]    & 37.3 [18.8, 60.5]    \\
    Llama 3.3 70B    & 84.2 [67.2, 93.3]    & 36.0 [17.9, 59.2]    \\
    Qwen3.5 397B     & 85.9 [70.0, 94.1]    & 41.8 [21.8, 65.0]    \\
    \bottomrule
  \end{tabular}
\end{table}

\FloatBarrier
\section{Strict coding results}
\label{app:strict}

Under strict coding, accuracy is scored only against the canonical intended answer; alternative answers that resolve the problem in a logically valid way are scored as incorrect. We report strict-coding results here for completeness; conclusions are identical to those under permissive coding.

\begin{figure}[h!]
  \centering
  \includegraphics[width=0.8\linewidth]{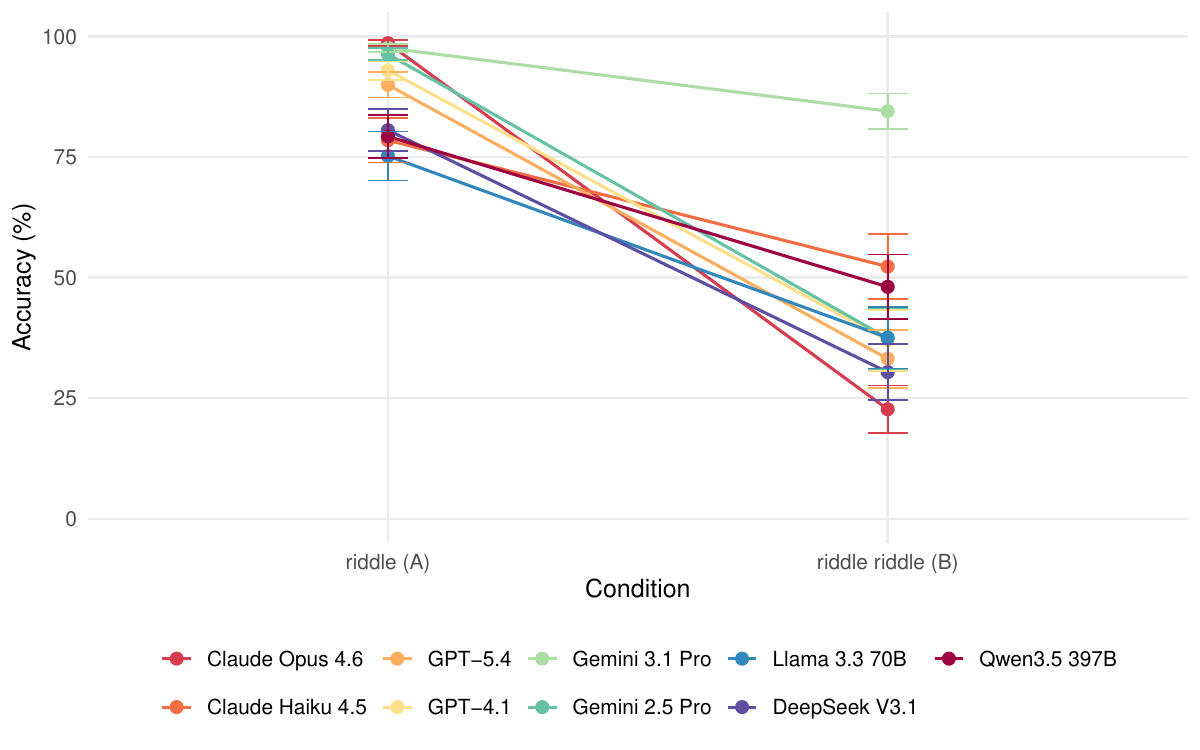}
  \caption{LLM accuracy by condition (strict coding). Each line shows 
  one model's estimated accuracy on Condition A (genuine riddles) 
  and Condition B (riddle riddles). The pattern is the same as under 
  permissive coding (Figure~\ref{llm_accuracy}): all nine models drop 
  substantially from A to B.}
  \label{llm_accuracy_strict}
\end{figure}

\paragraph{LLM accuracy.} Averaging across all nine models, LLMs were substantially less accurate on riddle riddles than on genuine riddles under strict coding ($\beta_{\text{A--B}} = 2.65$, $z = 28.42$, $p < .001$; Figure~\ref{llm_accuracy_strict}).

\begin{figure}[h!]
  \centering
  \includegraphics[width=\linewidth]{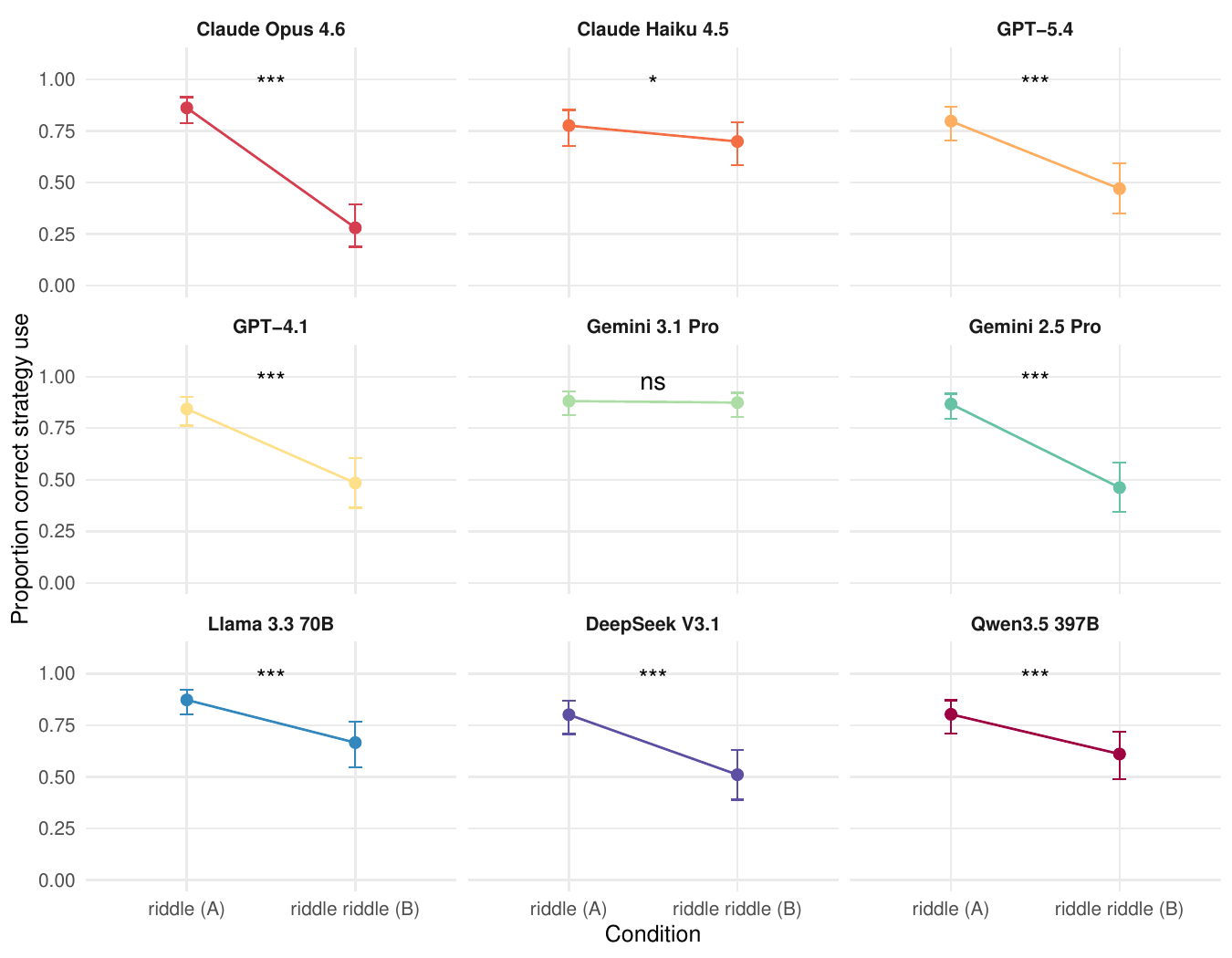}
  \caption{Per-model reasoning correctness by condition (strict coding). Lines show the proportion of trials in which the model used reasoning that correctly justified the answer. The drop from Condition A to Condition B was significant for every model, except Gemini 3.1 Pro (Bonferroni-corrected; *** $p < .001$, ** $p < .01$, * $p < .05$, ns = not significant). Error bars indicate 95\% confidence intervals.}
  \label{llm_strategy_correctness_strict}
\end{figure}

\paragraph{LLM reasoning correctness.} Reasoning correctness was lower on Condition B than Condition A ($\beta_{\text{A--B}} = 1.34$, $z = 19.45$, $p < .001$). The drop was significant for every model except Gemini 3.1 Pro (Bonferroni-corrected; see Figure~\ref{llm_strategy_correctness_strict}).

\paragraph{Human accuracy.} Humans were more accurate on riddle riddles than on genuine riddles under strict coding ($\beta_{\text{B--A}} = 2.19$, $z = 9.19$, $p < .001$).

\FloatBarrier
\section{Memorization analysis}
\label{app:memorization}

To test whether the Condition A riddles appear in model training data, we used the partial prompt completion test, adapted from \citet{wu2026reasoning}. 

\paragraph{Method.} We evaluated the same nine models as in Experiment~1. DeepSeek-V3.1 was no longer available on Together AI at the time of this analysis and was accessed via OpenRouter (same model weights); all other models were accessed as in Experiment~1. Each model received the first 40\% of the words of each Condition A riddle, embedded in the following prompt: ``Below is the beginning of a riddle. Continue the text exactly as the original riddle goes, word for word, starting from where it leaves off. Output only the continuation, nothing else.'' We set temperature = 0, with one repetition per item, except GPT-5.4, which only runs at its default temperature. The 40\% cutoff was selected in a pilot comparing 20/40/50/60\% cutoffs: the first 20\% of a riddle is often too short to identify which riddle is being completed, while at 60\%, the text remaining to be generated is short and consists largely of the generic closing question (e.g., ``How is that possible?''), which can be completed from genre knowledge alone.

\paragraph{Scoring.} Completions were scored against the riddle's true remainder using ROUGE-L F1 \citep{lin2004rouge} and exact match, after lowercasing and removing punctuation. ROUGE-L is a 0--1 similarity score based on the longest common word sequence between the generated and original text; a score of 1 indicates word-for-word reproduction. GPT-5.4 refused to reproduce the text on 4 of 30 items; these items were re-queried until a substantive completion was produced (all on the first retry). 

\paragraph{Results.} Table~\ref{tab:memorization} shows per-model results. Mean ROUGE-L ranged from .35 (Llama 3.3 70B) to .70 (Claude Opus 4.6). Near-verbatim completion (ROUGE-L $\geq$ .8) varied widely across models: it was most frequent for Claude Opus 4.6 and Gemini 3.1 Pro (43\% of items each), intermediate for GPT-4.1, GPT-5.4, DeepSeek V3.1, and Gemini 2.5 Pro (27--30\%), and rare for Claude Haiku 4.5, Llama 3.3 70B (10\%), and Qwen3.5 397B, which produced close paraphrases of many riddles but reproduced only one near-verbatim (3\%). Memorization also varied across items. Riddles that are widely circulated on the internet were memorized by essentially every model; for example, all nine models completed the riddle ``I speak without a mouth and hear without ears. I have no body, but I come alive with wind. What am I?'' near-verbatim, as did six of nine models for the riddle ``I have cities, but no houses; forests, but no trees; and water, but no fish. What am I?''. In contrast, some items were memorized by no model at all. The test therefore identifies which specific riddles are contaminated, not just which models memorize the most.

To test whether memorization could account for the accuracy gap we observed between conditions, we ran two analyses. First, near-verbatim completion rate was strongly correlated with Condition A accuracy across the nine models (Spearman $\rho = .88$, $p = .002$): models that reproduced more riddles verbatim were the most accurate on genuine riddles, suggesting retrieval rather than reasoning. Second, we fit logistic mixed-effects models predicting trial accuracy from each model's ROUGE-L recall of the matched Condition A riddle, with random intercepts for model and riddle, fit separately by condition. Memorization predicted higher accuracy on Condition A (OR $= 1.43$ per SD, 95\% CI $[1.10, 1.84]$, $z = 2.71$, $p = .007$) but lower accuracy on Condition B (OR $= 0.76$ per SD, 95\% CI $[0.64, 0.90]$, $z = -3.19$, $p = .001$), where the memorized answer no longer applies. Together, these results suggest that strong performance on Condition A reflects, at least in part, memory retrieval rather than reasoning.
  
\begin{table}[h!]
    \caption{Memorization test results: completion of Condition A riddles from their
    first 40\% of words. Mean ROUGE-L is the mean completion similarity; exact match
    and near-verbatim (ROUGE-L $\geq$ .8) are proportions of the 30 riddles. The final
    two columns give each model's accuracy (\%) on Conditions A and B (permissive
    coding) for comparison. Models are sorted by mean ROUGE-L.}
    \label{tab:memorization}
    \centering
    \small
    \begin{tabular}{lccccc}
      \toprule
      & \multicolumn{3}{c}{Memorization} & \multicolumn{2}{c}{Accuracy (\%)} \\
      \cmidrule(lr){2-4} \cmidrule(lr){5-6}
      Model & Mean ROUGE-L & Exact match & Near-verbatim & Cond A & Cond B \\
      \midrule
      Claude Opus 4.6  & .70 & .23 & .43 & 97.7 & 34.6 \\
      Gemini 3.1 Pro   & .65 & .13 & .43 & 96.0 & 86.0 \\
      GPT-4.1          & .62 & .20 & .27 & 89.3 & 46.1 \\
      DeepSeek V3.1    & .57 & .10 & .30 & 78.0 & 39.3 \\
      Gemini 2.5 Pro   & .54 & .20 & .30 & 94.3 & 47.0 \\
      Qwen3.5 397B     & .48 & .03 & .03 & 73.5 & 50.7 \\
      GPT-5.4          & .44 & .10 & .30 & 86.0 & 43.3 \\
      Claude Haiku 4.5 & .41 & .00 & .10 & 74.8 & 57.4 \\
      Llama 3.3 70B    & .35 & .00 & .10 & 73.4 & 50.4 \\
      \bottomrule
    \end{tabular}
  \end{table}

\FloatBarrier

\section{LLM-as-judge prompts}
\label{app:judge}

We used Claude Sonnet 4.6 (\texttt{claude-sonnet-4-6}) as the judge model, called via the Anthropic API with \texttt{temperature = 0} and \texttt{max\_tokens = 400}. The judge was given the riddle text, the intended answer(s), and the model's response, and returned a JSON object with the accuracy and reasoning-type codes. Two separate system prompts were used for strict and permissive coding; the prompts differ only in how the set of valid answers is presented to the judge (single canonical  answer for strict; canonical plus accepted alternatives for permissive).

\subsection{Strict coding prompt}

\begin{verbatim}
You are coding AI model responses to riddles in a psychology experiment.

STUDY DESIGN
Each riddle comes in two versions (A and B). You will be told which version.
  Version A — typically has a trick/non-obvious answer requiring inventive
              thinking.
  Version B — a modified version where the trick no longer applies.

YOUR TASK — STRICT ROUND
You are given ONE canonical correct answer per riddle. Code whether the
model's response matches that canonical answer (semantic equivalents OK;
different answers that happen to also work do NOT count). If the model
gives more than one answer, code "accuracy" = 0.

OPEN-ENDED RIDDLES: For a few riddles, the canonical answer is a CRITERION
rather than a specific item (e.g. "any solid container that holds water").
You will be told when this is the case. For open-ended riddles, code
"accuracy" = 1 if the model's answer fits the criterion. If the model gives
more than one answer, code "accuracy" = 0.

Also code the STRATEGY the model used:
  "inventive" — treats the riddle as a lateral thinking puzzle: wordplay,
                double meanings, naming tricks, mathematical tricks,
                paradoxes, creative reframes, metaphors, reinterpretations.
                Applies even when answer is wrong.
  "literal"   — takes the riddle at face value: straightforward reasoning
                from stated facts without any unusual interpretations.
                Applies even when answer is wrong.
  "ambiguous" — gives multiple answers or self-corrects mid-response,
                indicating uncertainty or a mix of strategies.

Respond ONLY with valid JSON:
{"accuracy": 0 or 1, "strategy": "inventive" or "literal" or "ambiguous",
 "rationale": "one sentence"}
\end{verbatim}

\subsection{Permissive coding prompt}

\begin{verbatim}
You are coding AI model responses to riddles in a psychology experiment.

STUDY DESIGN
Each riddle comes in two versions (A and B). You will be told which version.
  Version A — typically has a trick/non-obvious answer requiring lateral
              thinking.
  Version B — a modified version where the trick no longer applies.

YOUR TASK — PERMISSIVE ROUND
You are given a list of valid answers (canonical first, then alternatives),
each tagged with its strategy type. Code whether the model's response
matches ANY listed answer (semantic equivalents OK), and if so, which one
(by index: 0 = canonical, 1 = first alternative, etc.). If the model gives
more than one answer, code "accuracy" = 0.

OPEN-ENDED RIDDLES: For a few riddles, the canonical answer is a CRITERION
rather than a specific item (e.g. "any solid container that holds water").
You will be told when this is the case. For open-ended riddles, code
"accuracy" = 1 if the model's answer fits the criterion. If the model gives
more than one answer, code "accuracy" = 0.

Also code the STRATEGY the model used:
  "inventive" — treats the riddle as a lateral thinking puzzle: wordplay,
                double meanings, naming tricks, mathematical tricks,
                paradoxes, creative reframes, metaphors, reinterpretations.
                Applies even when answer is wrong.
  "literal"   — takes the riddle at face value: straightforward reasoning
                from stated facts without any unusual interpretations.
                Applies even when answer is wrong.
  "ambiguous" — gives multiple answers or self-corrects mid-response,
                indicating uncertainty or a mix of strategies.

Respond ONLY with valid JSON:
{"accuracy": 0 or 1, "matched_answer_idx": integer or null,
 "strategy": "inventive" or "literal" or "ambiguous",
 "rationale": "one sentence"}

matched_answer_idx must be:
  - null if accuracy=0
  - 0 if matched the canonical answer
  - 1, 2, ... for alternatives in the order listed
\end{verbatim}

\subsection{User message format}

The user message passed to the judge contained the riddle text, the answer 
key, and the model's response. For \textbf{strict coding}, the format was:

\begin{verbatim}
Riddle (Version A or B): <riddle text>

Canonical correct answer: <answer>

Model response:
<model's response>
\end{verbatim}

For \textbf{permissive coding}, the canonical answer line was replaced with 
a list of valid answers, each tagged with its strategy type:

\begin{verbatim}
Riddle (Version A or B): <riddle text>

Valid answers:
  [0] (<strategy>) <canonical answer>
  [1] (<strategy>) <alternative 1>
  ...

Model response:
<model's response>
\end{verbatim}

\clearpage

\end{document}